\pdfoutput=1
\documentclass[11pt]{article}

\usepackage{acl}
\usepackage{times}
\usepackage{latexsym}

\usepackage{float}
\usepackage{graphicx}
\usepackage{subfigure}
\usepackage{caption}
\usepackage{amsmath,amssymb}
\usepackage{mathtools}
\usepackage{booktabs}
\usepackage{multirow}
\usepackage{siunitx}
\usepackage{adjustbox}
\usepackage{pifont}

\usepackage[T1]{fontenc}
\usepackage[utf8]{inputenc}

\usepackage{microtype}
\usepackage{tikz}

\newcommand{\cmark}{\ding{51}}%
\newcommand{\xmark}{\ding{55}}%
\usepackage{physics}
\usepackage{comment}

\usepackage[normalem]{ulem}
\newcommand{\tbc}[1]{\textcolor{black}{#1}}

\newcommand{\hrupdate}[1]{\textcolor[rgb]{0.00,0.00,0.00}{#1}}
\title{You Don't Know My Favorite Color: Preventing Dialogue Representations from Revealing Speakers' Private Personas}

\author{{\bf Haoran Li}$^1$, {\bf Yangqiu Song}$^{1}$, {\bf Lixin Fan}$^2$\\
$^{1}$Dept. of CSE, Hong Kong University of Science and Technology\\
$^{2}$AI Group, WeBank \\
 \texttt{hlibt@connect.ust.hk},
 \texttt{yqsong@cse.ust.hk}, \\
 \texttt{lixinfan@webank.com}
\\
}
\DeclareMathOperator{\E}{\mathbb{E}}

\begin{document}
\maketitle
\begin{abstract}
Social chatbots, also known as chit-chat chatbots, evolve rapidly with large pretrained 
language models.
Despite the huge progress, privacy concerns have arisen recently: training data of large language models can be extracted via model inversion attacks.
%\yq{\sout{In addition} On the other hand} \yqc{I don't quite get the logic here when saying ``In addition''}
On the other hand, the datasets used for training chatbots contain many private conversations between two individuals.
In this work, we further investigate the privacy leakage of the hidden states of chatbots trained by %casual language modeling \yqc{what is casual LM?},
language modeling which has not been well studied yet.
We show that speakers' personas can be inferred through a simple neural network with high accuracy.
% can be more clear, name the MI LOSS and KL Loss
%To prevent such overlearning issues \yqc{change to ``persona leakage issues'' or simply ``To this end''}
To this end, we propose effective defense objectives to protect persona leakage from hidden states.
We conduct extensive experiments to demonstrate that our proposed defense objectives can greatly reduce the attack accuracy from 37.6\% to 0.5\%.
Meanwhile, the proposed objectives preserve language models' powerful generation ability. 

%With consistent personas and large pretrained casual language models, chit-chat chatbots, also known as social chatbots.
%Despite chatbots' 
\end{abstract}

\section{Introduction}
Social chatbots have been widely used to benefit many applications from answering factual questions to showing emotional companionship.
With recent progress in large pretrained language models~\cite{radford-2019-language,Yang-2019-XLnet}, some attempts ~\cite{Wolf-2019-transfer,zhang-2020-dialogpt,ham-etal-2020-end,Shen-2021-dialogXL,sevegnani-etal-2021-otters,gu-etal-2021-pral} are made to build chatbots based on large generative language models  (LMs).
%like GPT-2 ~\cite{radford-2019-language} and XLNet ~\cite{Yang-2019-XLnet}.
To train such LM-based chatbots, private conversations are collected.
Unfortunately, large language models tend to memorize training data and some private data can be recovered from models ~\cite{Pan-2020-Privacy,carlini-2021-extracting}.
%via black-box training data extraction attacks ~\cite{carlini-2021-extracting}.
%%% NEED transition!!!
%Recent studies propose differential privacy ~\cite{DP-book} and unlikelihood training ~\cite{Welleck-2019-neural} to mitigate the memorization issue. %Yangqiu: I deleted this sentence because it breaks the logic.
Besides such memorization problems, 
``overlearning'' on simple training objectives can reveal sensitive attributes indirectly related to the learning task~\cite{song-2020-overlearning}.
%hidden representations of machine learning models with simple objectives \cite{song-2020-overlearning} may reveal sensitive attributes of inputs.
%However, few studies consider the overlearning problem with LM.
%Hence, we raise the following research questions.
LM-based social chatbots essentially inherit the privacy issues of general LMs and the overlearning problem. 
% \yq{First}, without revealing the training data, are LM-based chatbots private enough for serving users?
% If the answer is no, then to what extend the privacy may be violated and how to address such privacy violation?

% \begin{table}[t]
% \small
% \begin{tabular}{p{0.9\linewidth}}%{l}
% %Dialog Context \\
% \hline
% \textbf{Dialog Context} \\
% \textbf{Speaker A}: Hi! \\
% \textbf{Speaker B}: Hello! How are you today? \\
% \textbf{Speaker A}: I am good, thank you. How are you?\\
% \textbf{Speaker B}: Great, thanks! My children and I were just about to watch Game of Thrones.\\
% \textbf{Speaker A}: Nice! How old are your children?\\
% \textbf{Speaker B}: I have four that range in age from 10 to 21, you?\\
% \textbf{Speaker A}: I do not have children at the moment.\\
% \textbf{Speaker B}: That just means you get to keep all the popcorn for yourself.\\
% \textbf{Speaker A}: And cheetos at the moment!\\
% %\textbf{Speaker B}: \\
% %\textbf{Speaker A}: \\
% \hline
% \end{tabular}
% %Dialog Context \\
% %\hline
% %Speaker A: \\
% \captionsetup{type=figure}
% \caption{One dialog sample from the PersonaChat dataset.}
% \label{fig:dialog sample}
% \vspace{-0.2in}
% \end{table}

\begin{figure*}[t]
\centering
\makebox[0pt]{
\includegraphics[width=1\textwidth]{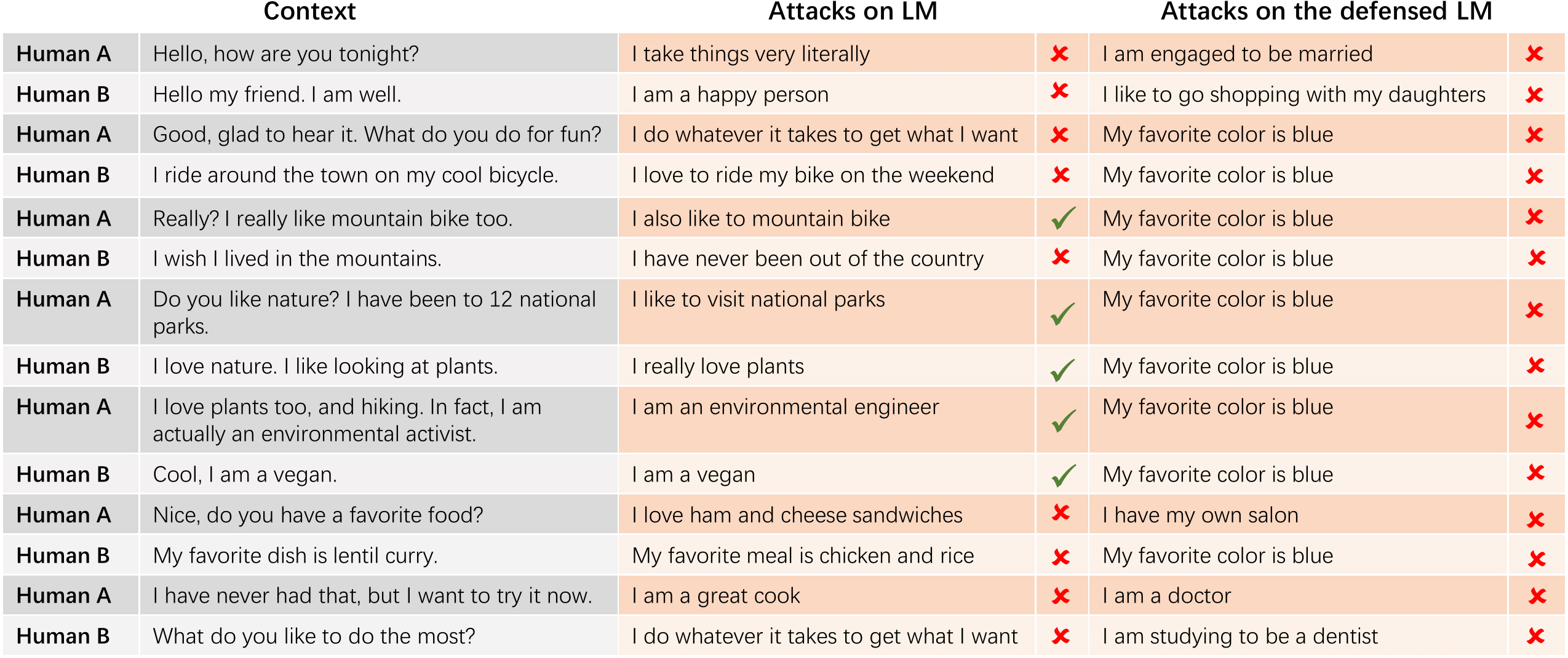}
}
\vspace{-0.05in}
\caption{Black-box persona inference attacks (over 4,332 personas) on a dialog.
Every representation of the utterance, which is based on the last hidden state of GPT-2, is attacked without defense (column of ``Attacks on LM'') and with defense (column of ``Attacks on the defensed LM''). If the model can predict the persona of the speaker based on the observed representation, then we regard it as a successful attack; otherwise, unsuccessful. In practice, when deploying a model, a robust model which will reveal nothing of the encoded utterances is expected. %\yqc{Also mention it's a black-box attack model.}
}
\label{fig:atk-full-sample}
\vspace{-0.15in}
\end{figure*}

%The answer to the first question is indeed no.
% For example, besides explicitly extracting training data from language models, implicit attributes can also be inferred for dialog systems.
% From the conversation shown in Figure ~\ref{fig:dialog sample}, we can explicitly know that B has 4 young children and they like to watch Game of Thrones while A is childless.
% Moreover, we may implicitly infer that both A and B are humorous and friendly through their last utterances.

For example, as Figure~\ref{fig:atk-full-sample} shows, when using a fine-tuned GPT-2 as the encoder and decoder of an LM-based social chatbot, if the learned representation of each utterance can be obtained by an adversary, then the adversary can build a classifier to predict the persona information based on the representation. 
As shown by the example, for five out of 14 utterances, the attacker can successfully predict the persona, which can be harmful if the users (speakers of the utterances) do not prefer to reveal the persona information.
Thus, in practice, when deploying such kinds of chatbots in real applications, we should first make sure that no private information can be leaked by the models.

% % need to put some models here
% Recent \yq{research studies} propose several model architectures ~\cite{Tigunova-2019-ListeningBT,Wu-2020-GettingTK} to infer speakers' attributes based on their utterances during conversation. They mostly consider constructing new models to capture implicit attributes and neglect the fact that neural dialog models can be directly used to infer speakers' attributes.

To systematically study the privacy issues in LM-based social chatbots, there are several challenges.
First, there is no existing data that can be used to quantify how much private information is revealed by an LM.
%\hrupdate{First, there is no standalone dataset to evaluate privacy leakage of training data in a LM ~\cite{carlini-2021-extracting}.}
Second, there has been no existing work showing how to attack utterance-level representations to obtain sensitive information.
%an LM-based dialogue system and to what extent an attack can obtain sensitive information. \yqc{Go through the paper to distinguish from ML attack/defense.}
Third, there has been no existing LM-based chatbot that can defend against persona inference attacks, and no study shows how to protect both known and unknown persona attributes.

%%%%\yqc{Can you revise the approach part by addressing challenges? First, we build a new dataset, second, we attack, third, we defend.}

In this paper, to address the above challenges, we use the fine-tuned GPT-2 as our chatbot.
We first collect a dataset by aligning personas with corresponding utterances in PersonaChat dataset ~\cite{zhang-etal-2018-personalizing}.
Then we show that ``overlearning'' can happen for LM-based chatbots to reveal personas of speakers.
We build a single external multi-layer perception (MLP) attacker model to perform black-box persona inference attacks on the utterance-level embeddings.
With no access to parameters of the chatbot, the attacker model can infer speakers' personas with 37.59\% accuracy over 4,332 personas.
%\yqc{At this point, we need to show that it's black-box attack, which does not know GPT-2's parameters but only the final embedding of an untterance.} which is capable of inferring speakers' personas with 37.59\% accuracy over 4,332 personas.
The high accuracy of the attacker model implies that the utterance-level embeddings have potential vulnerabilities to reveal speakers' private persona attributes. 
Thus, it is necessary to improve training algorithms to address such overlearning issues. 
Finally, we apply defense learning strategies on the GPT-2 to prevent such black-box attacks.
We combine proposed \tbc{KL divergence loss (KL loss) with mutual information loss (MI loss)} \cite{song-2019-learning} as additional defense objectives to train the GPT-2 and decrease the attacker's persona inference accuracy to 0.53\%.
Our contributions can be summarized as follows:\footnote{Code is publicly available at \url{https://github.com/HKUST-KnowComp/Persona_leakage_and_defense_in_GPT-2}.}

1): To the best of our knowledge, we are the first to disclose and analyze the persona inference attack for LM-based chatbots and treat it as a privacy risk.
  % need better names for two losses
  
2): We propose an  effective defensive training algorithm to prevent dialog representations from leaking personas of the corresponding speakers by uniform distribution approximation and mutual information minimization.

3): We conduct extensive experiments to quantify both privacy and utility of proposed defense mechanisms.
Besides solving the persona leakage issue, the proposed training algorithm  has nearly no negative influence on utility. 

%For example, it is possible to obtain individuals' names, phone number and 

%Companies implement their commercial chatbots for factual question answering and customer services, and  

\section{Related Work}
Language models trained on private data suffer privacy risks of revealing sensitive information. 
Previous researches mainly considered black-box attacks that assumed attackers only had access to inputs and outputs of language models.
~\citet{carlini-2021-extracting} performed black-box model inversion attack on GPT-2 through descriptive prompts with beam search.
~\citet{lehman-2021-bert} examined BERT pretrained on Electronic Health Records via blank filling and model probing to recover Personal Health Information.
Furthermore, given black-box access to  a language model's pre-train and fine-tune stages, ~\citet{zanella-bguelin-2020-analyzing} showed that sensitive sequences of the fine-tuning dataset can be extracted.
For the distributed client-server setup, \citet{malekzadeh-2021-honest} considered the sensitive attribute leakage from the server side with honest-but-curious (HBC) classifiers.

What is worse, for an LM-based chatbot, its training conversations are prone to include more private attributes than other commonly-used corpora for language modeling like BooksCorpus ~\cite{Zhu-2015-Aligning} and Wikipedia.
~\citet{Tigunova-2019-ListeningBT} proposed Hidden Attribute Model (HAM) to extract professions and genders of speakers from various dialog datasets.
~\citet{Wu-2020-GettingTK} further applied Attribute Extractor to generate speakers' attribute triplets flexibly and suggested downstream tasks based on the triplets.
~\citet{Pan-2020-Privacy} exploited embeddings of language models to recover inputs' digits and keywords.
Though the setup of this work is similar to ours, they merely consider simple cases of data recovery with given rules and suffer great utility degradation to obtain optimal defense performance.
%%%% Need transition
For our work, there is no fixed pattern or rule for the model input.
Instead of finding keywords or recovering digits, we aim to infer more complicated private attributes from such embeddings. Moreover, our proposed defenses have almost no influence on the utility.

\section{Attacking on Language Models}
In this section, we illustrate black-box persona inference attacks on GPT-2 and our defense strategies.
In Section~\ref{sec:3-1}, we first give the problem formulation. 
Then we describe the attack in Section~\ref{sec:3-2}.
%Lastly, we comprehensively explain our proposed defense strategies in Section~\ref{sec:3-3}.

%%%%%%%%%%   Model  Figure
\begin{figure*}[h]
\centering
\includegraphics[width=1\textwidth]{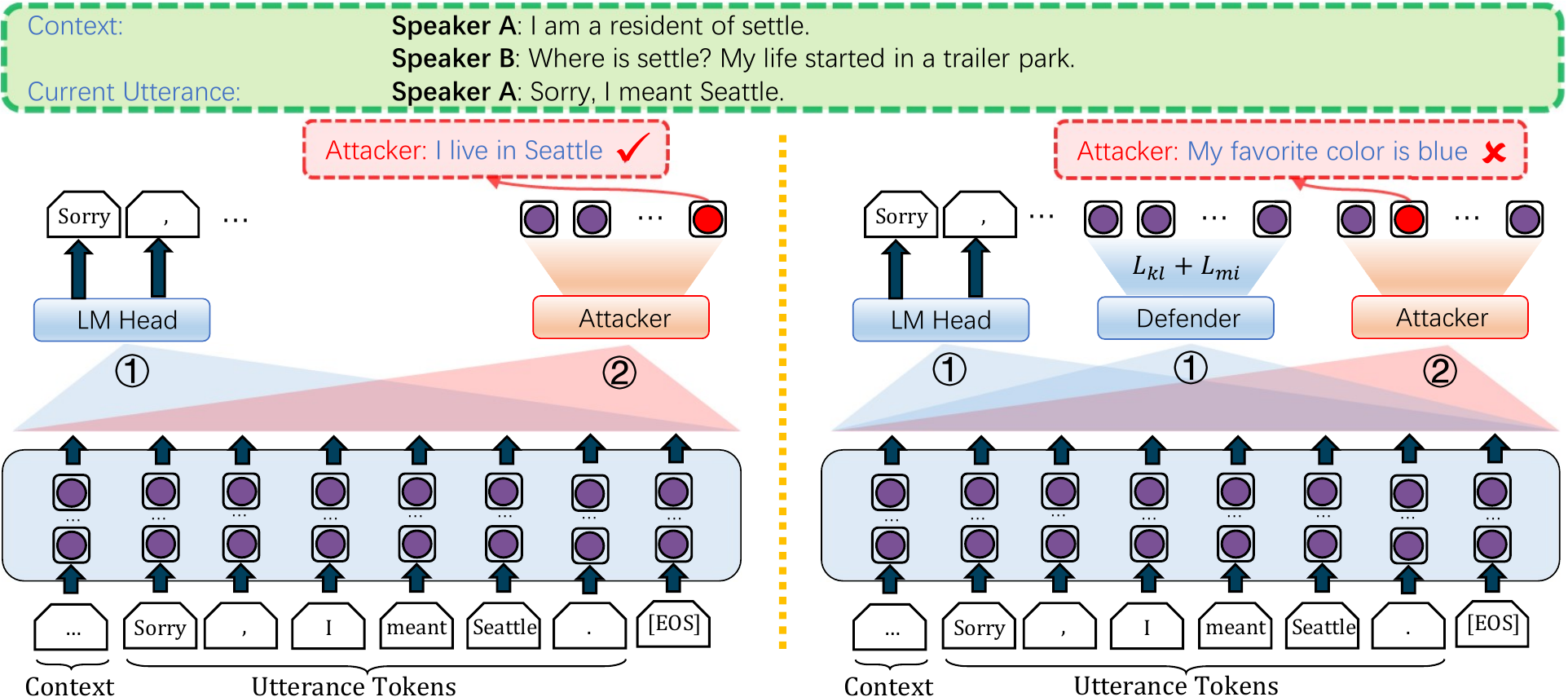}
%\vspace{-0.1in}
\caption{
Scenarios for attacks without defense (left) and with defense (right). 
%\textcircled{\raisebox{-0.9pt}{1}} indicates GPT-2's training stage before \textcircled{\raisebox{-0.9pt}{2}}.
%\textcircled{\raisebox{-0.9pt}{2}} refers to the attacking stage where \textcircled{\raisebox{-0.9pt}{1}} is finished and parameters of GPT-2 are all frozen.
The GPT-2's training stage is marked by \textcircled{\raisebox{-0.9pt}{1}} and the attacking stage is marked by \textcircled{\raisebox{-0.9pt}{2}}.
Both language modeling and defender objectives are jointly trained for the defense to optimize the GPT-2 model.
After GPT-2's training stage \textcircled{\raisebox{-0.9pt}{1}} is finished, parameters of GPT-2 are all frozen and then the attacking stage \textcircled{\raisebox{-0.9pt}{2}} starts.
The defender shares the same architecture as the attacker and uses $L_{kl}$ with $L_{mi}$ as defense objectives.
}
\label{fig:model}
\vspace{-0.1in}
\end{figure*}

\begin{comment}
\begin{figure*}[h]
\centering
\setlength{\abovecaptionskip}{-0.0cm}
\subfigure[Attack on GPT-2 without defense.]
{\label{fig:attack_model}
\begin{minipage}[t]{0.49\linewidth}
%\vspace{-0.05in}
\includegraphics[scale=0.48]{imgs/attack_model_crop.pdf}
\end{minipage}%
%\vspace{-0.05in}
}
\subfigure[Attack on GPT-2 with defense.]
{\label{fig:defense_model}
\begin{minipage}[t]{0.49\linewidth}
%\vspace{-0.05in}
\includegraphics[scale=0.48]{imgs/defense_model_crop.pdf}
\end{minipage}%
%\vspace{-0.05in}
}
\centering
\caption{Overall pipeline for attack and defense scenarios. 
The orange dashed arrows indicate objectives of the GPT-2 and the \tbc{blue arrows} are used by the adversary.
For both cases, the GPT-2 is trained at first and its parameters are all frozen after training.
Then the adversary trains its attacker model based on embeddings of the GPT-2.
}\label{fig:model}
\end{figure*}
\end{comment}
%%%%%%%%%% need to change color of attacker arrow, also add stage 1 and stage 2.

\subsection{Problem Formulation}
\label{sec:3-1}
We assume that there is a GPT-2 based chatbot $f$ pretrained on private conversations $D$.
Only language modeling is used to train the chatbot:
\begin{equation}
\label{eqn:LM}
\small
L_{f} (u;\theta_f) =
-\sum_{i=1}^{|u|} \log(\text{Pr}(w_i|c,w_0,w_1,...,w_{i-1})),
\end{equation}
%\yqc{$\sum_{i=1}^{|U|}$ to be $\sum_{i=1}^{|\vb{U}|}$?}
%where the negative log likelihood with softmax is applied to estimate the probability distribution $\text{Pr}(w_i|c,w_0,w_1,...,w_{i-1})$ \yqc{Pr should be a function of $\theta_f$.}
%$L_{f}$ is casual language modeling loss 
%for target dialog language model $f$ with given utterance $u=\{w_0,w_1,...,w_{|u|-1} \}$ and previous context $c$ in $D$.
where $f$ refers to the LM-based chatbot with given utterance $u=\{w_0,w_1,...,w_{|u|-1}\}$ and previous context $c$.
An adversary owns one external annotated dialog dataset $
D_a  = \{(\vb{U}_1,\vb{s}_1),(\vb{U}_2,\vb{s}_2), ... , (\vb{U}_n,\vb{s}_n)\} 
$
with $n$ conversations where $\vb{U}_i$ indicates a list of utterances $\{u_{i1},u_{i2},...,u_{in_i}\}$ of $i$-th conversation and $\vb{s}_i$ corresponds to a list of sensitive personas $\{s_{i1},s_{i2},...,s_{in_i} \}$ for corresponding utterance.
Each persona $s_{kj}$ is an integer that can be mapped to its persona according to a predefined dictionary and $0 \leq s_{kj} \leq C-1$ where $C$ is the total number of predefined persona attributes.
%The adversary has no access to original training data $D$ and hence there is no distributional similarity guarantee between $D_a$ and $D$.
The goal of the adversary is to infer speakers' personas $s$ from utterances' embeddings $f(u)$ where $u$ and $s$ refer to any utterance and its persona label. 
%Moreover, $u$ and $s$ are not in $D_a$ and there is no fixed pattern or rule for $u$.

%%% \input{fig-model}

\subsection{Black-box Persona Inference Attack}
\label{sec:3-2}
The persona inference attack can be viewed as a supervised classification task.
For the black-box attack setup, the adversary can only query the target dialog model $f$ with access to embeddings of adversary's inputs and cannot access or modify model parameters $\theta_{f}$.
As shown in the left part of Figure~\ref{fig:model}, the adversary tries to build its attacker model $\mathcal{A}$ with its external data $D_{a}$ and dialog model $f$.
The persona predictor's output $\mathcal{A}(f(u))$  is the estimated probability distribution over $C$ persona attributes.
Its loss function $L_{\mathcal{A}}$  exploits cross-entropy between the predicted distribution and ground truth distribution that can be formulated as:  
\begin{equation}
\label{eqn:black-box attack}
\small
% L_{\mathcal{A}} = \sum_{k=1}^{n} \sum_{j=1}^{n_k} - \mathcal{A}(f(u_{kj}))_{y_{kj}} + \log \sum_{i=0}^{C-1} \mathcal{A}(f(u_{kj}))_i
L_{\mathcal{A}} (u_{kj},s_{kj};\theta_{\mathcal{A}}) = 
\text{CE}(
\mathcal{A}(f(u_{kj})) , s_{kj}
),
\end{equation}
where CE refers to cross-entropy loss between persona label $s_{kj}$ and $\mathcal{A}(f(u_{kj}))$.

A well-performed persona predictor $\mathcal{A}$ can cause great privacy threats.
For machine learning as a service (MLaaS),  $\mathcal{A}$ can be applied to perform a  man-in-the-middle attack on the application programming interfaces.
Moreover, even if the raw data are protected and the  transmission channel is secure, a curious service provider can train its attacker  $\mathcal{A}$ to collect personas of service users.

%\subsection{Defense Learning Strategies}
\section{Defense Learning Strategies}
\label{sec:3-3}
The LM training objective \hrupdate{in Equation~\ref{eqn:LM}} only considers the utility of chatbots.
In later \tbc{experiment sections}, we show that LM brings severe overlearning issues. 
%%%%% removed by suggestions from reviewers
%%%%%To avoid black-box persona inference attacks, more constraints on privacy should be added for training LM-based chatbots.
Ideally, to achieve an optimal privacy-preserving chatbot against persona inference attacks, the probability distribution of the attacker model $\mathcal{A}$ should be close to the uniform distribution.
That is, the adversary cannot improve its inference accuracy from posterior estimation $\mathcal{A}(f(u))$ and the accuracy is no better than making random guesses on the persona attributes.
Moreover, the constraints on privacy should have minor degradation on the utility to maintain the strong generation ability of chatbots.

Following the intuition that the adversary cannot obtain better results than a random guess, in Section~\ref{sec: KL Loss}, we propose \tbc{KL loss}  that aims to flatten the persona predictor's estimated distribution.
Based on minimizing the mutual information between hidden states $f(u)$ of chatbots and private persona attributes $s$, we propose \tbc{MI loss} in Section~\ref{sec: MI Loss}.
Lastly, we show the overall training objective in Section~\ref{sec: Overall Objective}.
%For the utility, we still use Equation~\ref{eqn:LM} as the objective function. %\yqc{Can you check all equation references to be somthing like Eq.~(1)?} as the objective function.

%\subsubsection{KL Loss}
\subsection{KL Loss}
\label{sec: KL Loss}
\tbc{KL loss} aims to minimize the Kullback–Leibler divergence between $\mathcal{A}(f(u))$ and the uniform distribution.
It flattens the distribution of $\mathcal{A}(f(u))$ so that the adversary cannot gain any useful knowledge after training attacker model $\mathcal{A}$.
%After simplification, the KL divergence can be formulated as~\cite{mireshghallah-2021-privacy}:
The KL divergence between the uniform distribution and $\mathcal{A}(f(u))$ can be formulated as:
\begin{equation}
\label{eqn:KL Loss 0}
\small
%D_{KL}(u;\theta_{\mathcal{A}}) \propto
D_{KL} (\text{UNI} || \mathcal{A}(f(u))) =
-\frac{1}{C}\sum_{k=0}^{C-1}  \log (C\text{Pr}(k | f(u),\theta_{\mathcal{A}})),
\end{equation}
where UNI indicates the uniform distribution and $k$ indicates the $k$-th persona label \hrupdate{of $C$ labels}. 
For optimization, we can leave out constant terms \hrupdate{and the logarithm} ~\cite{mireshghallah-2021-privacy} to obtain the following loss function:
\begin{equation}
\label{eqn:KL Loss 1}
\small
L_{D}(u;\theta_{\mathcal{A}}) =
-\frac{1}{C}\sum_{k=0}^{C-1}  \text{Pr}(k | f(u),\theta_{\mathcal{A}}).
\end{equation}
However, from the perspective of defenders, they have no access to attacker model $\mathcal{A}$ and its parameters.
Instead, they can build their own persona predictor as a fake attacker. 
More specifically, they may mimic the adversary to annotate a dataset $D_{a}'$ and a persona predictor $\mathcal{A}_p$.
Then the \tbc{KL loss} becomes:
\begin{equation}
\label{eqn:KL Loss 2}
\small
L_{kl} (u;\theta_{\mathcal{A}_p},\theta_{f}) = 
-\frac{1}{C}\sum_{k=0}^{C-1}  \text{Pr}(k | f(u),\theta_{\mathcal{A}_p}),
\end{equation}
where parameters of the chatbot $\theta_{f}$ and the fake attacker $\theta_{\mathcal{A}_p}$ are updated via \tbc{KL loss}.
The intuition is to train the chatbot together with a fake attacker to prevent model overlearning by flattening the attacker model's distribution.

%\subsubsection{MI Loss}
\subsection{MI Loss}
\label{sec: MI Loss}
The privacy constraint requires that hidden representations should not reveal the persona attributes.
In other words, given any utterance $u$ and persona $s$ behind the utterance $u$, we want to minimize the mutual information between $f(u)$ and $s$:
\begin{equation}
\small
\label{eq:mi}
 \min\limits_{\theta_{f}}I(f(u);s).
    %\hat{\epsilon} = \alpha+\log(\frac{1}{\delta}),
\end{equation}
%%% how to cite?
Following the derivation in \citet{song-2019-learning} and \citet{Li-2020-TIPRDCTP}, the upper bound can be formulated as:
\begin{equation}
\small
\label{eq:mi_upper1}
 I(f(u);s) \leq \E_{q(f(u))} D_{KL} (q(s|f(u))||p(s)),
\end{equation}
where $p(s)$ can be any distribution for $s$,
$q(x)$ refers to probability distribution of model $f$ parameterized by $\theta_f$ and $f(u)$ is assumed to be sampled from the conditional distribution $q(f(u)|x,s)$.
However, $q(s|f(u))$ is hard to estimate.
Instead, we use $p_{\Psi}(s|f(u))$ to approximate $q(s|f(u))$ via minimizing their KL divergence and then we can obtain the following lower bound \cite{song-2019-learning}:
\begin{equation}
\small
\label{eq:mi_upper2}
\begin{multlined}
 \E_{q (f(u))} D_{KL} (q(s|f(u))||p(s)) \\
 \geq \E_{q (f(u))} [\log p_{\Psi}(s|f(u))-\log p(s)].
\end{multlined}
\end{equation}
%\yqc{Can you put a detailed derivation in the Appendix? Here, I think more intuitions should be added.}
Therefore, our objective in Equation \ref{eq:mi} can be formulated as an adversarial training objective:
\begin{equation}
\small
\label{eq:mi_upper3}
 \min\limits_{\theta_{f}} \max\limits_{\Psi}
 \E_{q (f(u))} [\log p_{\Psi}(s|f(u))-\log p(s)].
\end{equation}
$\log p(s)$ is independent of $f(u)$, and we may leave this term out in Equation \ref{eq:mi_upper3}: 
\begin{equation}
\small
\label{eq:mi_upper4}
 \min\limits_{\theta_{f}} \max\limits_{\Psi}
 \E_{q (f(u))} [\log p_{\Psi}(s|f(u))].
\end{equation}
Then, Equation \ref{eq:mi_upper4} illustrates an adversarial game between an adversary $p_{\Psi}$ who manages to infer $s$ from $f(u)$ and a defender who modifies $\theta_{f}$ to protect $s$ from persona inference attack.
%%%% hr added for more refs
Adversarial training is widely used to protect sensitive features in natural language processing \cite{elazar-goldberg-2018-adversarial,coavoux-etal-2018-privacy,li-etal-2018-towards}.
Using  the persona predictor model $\mathcal{A}_p$ with softmax activation to learn $p_{\Psi}$, we obtain the final objective for the defender:
\begin{equation}
\small
\label{eq:mi_upper5}
 \min\limits_{\theta_{\mathcal{A}_p}}%\theta_{f}
 \max\limits_{\theta_{f}}
 \text{CE}(
\mathcal{A}_p(f(u)) , s
).
\end{equation}
We can rewrite Equation \ref{eq:mi_upper5} into two losses:
$
\small
L_{mi1} (u_{kj},s_{kj};\theta_{\mathcal{A}_p}) = 
\text{CE}(
\mathcal{A}_p(f(u_{kj})) , s_{kj})
$
and 
$
\small
L_{mi2} (u_{kj},s_{kj};\theta_{f}) = 
-\text{CE}(
\mathcal{A}_p(f(u_{kj})) , s_{kj})
$
for the fake adversary and the chatbot respectively.
Then our MI loss can be formulated as:
\begin{equation}
\small
\label{eq:final_mi}
 L_{mi} = \lambda_0 L_{mi1} + L_{mi2},
\end{equation}
where $\lambda_0$ controls the ratio between two the fake attacker $\mathcal{A}_p$ and the defensed chatbot $f$.

%\subsubsection{Overall Objective}
\subsection{Overall Objective}
\label{sec: Overall Objective}
The right part of Figure \ref{fig:model} illustrates how the chatbot is trained to address the black-box attack.
%%% can add more explanation here
The loss function for the defender combines \tbc{KL loss, MI loss} and LM loss.
Notice that the fake adversary objective in \tbc{MI loss} violates \tbc{KL loss} which tries to make the distribution of $\mathcal{A}_p$ flatten.
Our proposed loss assigns more weights to the \tbc{KL loss}:
\begin{equation}
\label{eqn:overall loss}
\small
L = L_f + \lambda_1 L_{kl} + \lambda_2 L_{mi},
\end{equation}
where $\lambda_1$ and $\lambda_2$ are hyper-parameters and $\lambda_1 \geq 10\lambda_2$ to flatten the distribution of $\mathcal{A}_p$.
Though the chatbot trained with overall loss $L$ still cannot interfere training process of $\mathcal{A}$ during black-box attacks, $L$ aims to mitigate persona overlearning issues of $f$ to address such persona inference attacks.

\begin{comment}
\begin{equation}
\label{eqn:lmi1}
\small
L_{mi1} (u_{kj},s_{kj};\theta_{\mathcal{A}_p}) = 
\text{CE}(
\mathcal{A}(f(u_{kj})) , s_{kj}
),
\end{equation}
\begin{equation}
\label{eqn:lmi2}
\small
L_{mi2} (u_{kj},s_{kj};\theta_{f}) = 
-\text{CE}(
\mathcal{A}(f(u_{kj})) , s_{kj}
),
\end{equation}
\end{comment}
%That is, parameters of the chatbot $\theta_{f}$ aim to protect persona attributes and parameters of the persona predictor $\theta_{\mathcal{A}_p}$ tries to infer

%For grey-box attack, we follow previous grey-box adversarial attack setup~\cite{Xu-2021-GreyboxAA} that allows the adversary to update parameters of target model $f$ only during the training phase.
%Still, for inference stage, the adversary only has access to embeddings $f(u)$ and aims to conjecture on the persona attributes $y$.
%Intuitively, the adversary has higher attack success rate than black-box setup for persona predictor 
%Since parameters of target dialog model $\theta_{f}$ can be trained with persona predictor,  
%a higher persona inference rate is expected for the grey-box attack.

%\begin{equation}
%\label{eqn:grey-box}
%\small
%L_{G} = L_{\mathcal{A}} %(\theta_{\mathcal{A}};\theta_{f}) + L_{f},
%\end{equation}

\section{Experiments}
In this section, we conduct experiments to evaluate the performance of privacy and utility for the proposed defense learning strategies.
In Section \ref{exp:4.1}, we give our experimental settings in detail.
In Section \ref{exp:4.2}, we show the attacking performance with and without defense.
In Section \ref{exp:ablation}, we perform ablation study on defense objectives.
In Section \ref{exp:4.3}, we use automatic metrics to evaluate chatbots' utility.
We conduct various attack setups in Section \ref{exp:More Setups} and perform a case study in Section \ref{case}.

\subsection{Experimental Settings}
\label{exp:4.1}
\begin{table}
\centering
%\small
\resizebox{0.37 \textwidth}{!}{
\begin{tabular}{lr}
\hline
\textbf{Stat Type} & \textbf{Value} \\
\hline
Dialogs & 10,907  \\
Utterances (turns)  & 162,064 \\
Unique personas &  4,332  \\
Total personas & 98,056 \\
Labeled turns & 32,147 \\
Avg. turns per dialog & 14.86 \\
Avg. labeled turns per dialog & 2.95 \\
Avg. words per turn &  11.71  \\

\hline
\end{tabular}
}
\vspace{-0.1in}
\caption{\label{tab:dataset-table}
Statistics of the aligned dataset.
}
\vspace{-0.15in}
\end{table}

\begin{comment}
\begin{table}[t]
	\footnotesize
	\label{tab:dataset-table}
	\vspace{-0.1in}
	\resizebox{0.4 \textwidth}{!}{
       \begin{tabular}{ccc}
			\toprule
			\multicolumn{2}{c|}{Stat Type} & Value \\ \midrule
		    \multicolumn{2}{c|}{Dbpedia} & 14,085 \\
		    \multicolumn{2}{c|}{Dbpedia} & 14,085 \\
		    \bottomrule
	   \end{tabular}
	}\vspace{-0.1in}
	\caption{Statistics of the Knowledge Graphs.}
\end{table}
\end{comment}
\textbf{Dataset.}
To train the GPT-2 as our chatbot, we use the DialoGPT ~\cite{zhang-2020-dialogpt} pretrained on Reddit comment chains.
Then we use PersonaChat dataset ~\cite{zhang-etal-2018-personalizing} to fine-tune the GPT-2.
To obtain annotated dataset $D_a$ for the adversary, we align personas to corresponding utterances through positive (utterance,persona) pairs provided in Dialogue NLI ~\cite{welleck-etal-2019-dialogue} dataset.
For those utterances with no annotations, we assign label $-1$ to them.
We reshuffle the dataset to balance the label distribution among train/val/test datasets with the ratio of $8:1:1$. 
%and divide them with the ratio of $8:1:1$. 
%\yqc{I think you can simply keep it 8:1:1 for submission.}
We first let the attacker and defender share the same training data.
In later sections, we will separate the annotated data for the adversary and defender with no overlap. 
A summary statistics of $D_a$ is shown in Table~\ref{tab:dataset-table}.

\textbf{Attacker model.}
In our experiment, we use a 2-layer neural network with cross-entropy loss as the attacker model.
The attacker model exploits the final layer embedding of the last token ``<|endoftext|>'' from the GPT-2 as model input.
We also try other attacker model architectures (transformer block based attackers) and input embeddings (average of all embeddings in the final layer of GPT-2), but the attacking performance is worse than the 2-layer model mentioned above.

%%% detail, hyperparameters

\textbf{Evaluation Metrics.}
The evaluation metrics are based on privacy and utility.
For privacy, we use persona inference accuracy and weighted F1-score to evaluate the attacker's performance.
We also use Bayesian Privacy (BP)~\cite{gu-2021-federated} to quantify the attacker's privacy loss for the estimated persona distribution.
Top-k accuracy is reported in the Appendix.
For utility, we apply BERTScore ~\cite{bert-score}, Distinct ~\cite{li-2016-diversity}, BLEU ~\cite{Papineni-2002-BLEU} and perplexity (PPL) as evaluation metrics.
BERTScore and BLEU measure similarity between generated outputs and ground truth while Distinct (Dist) focuses on  diversity.
Perplexity shows the uncertainty when the LM model fits the data.

%%%%%%
%What exp to put?
%1): Result on privacy, before defense, after defense, (connect with overlearning), maybe i can also put albation here
%2): Generation quality, not much influence
%3): Zero shot setting
%4): 8 clusters with detailed distribution 

%%%%%%
\begin{table}
\begin{adjustbox}{width=0.9\columnwidth,center}
\centering
\small
  \begin{tabular}{lccc}
    \toprule
%    \multirow{1}{*}{Dataset}  \\
% $\downarrow$
    {}  & {Acc} & {F1} & {Max-Ratio}    \\
      \midrule
    
    Random Pred & 0 & 0 & {0.02}    \\
    
    \tbc{Best Guess} & 0.72 & 1.02e-3 & 100 \\
    
    LM & 37.59 & 3.65e-1 & 1.34   \\
    LM+KL+MI & 0.53 & 6.78e-5 & 81.87   \\
    \midrule
    
    LM+KL & 14.43 & 1.13e-1 & 10.60   \\
    %\multirow{7}{*}{}  &
    \tbc{LM+MI} & 0.53 & 5.57e-5 & 99.84   \\
    
    %LM+KL+MI & 0.53 & 6.78e-5 & 81.87   \\
    %\hline
    %Imbalance & 0.47  &  1.90e-3  & 94.06   \\

    \bottomrule
  \end{tabular}
  \end{adjustbox}
  \vspace{-0.05in}
  \caption{\label{tab:privacy}
Evaluation on the privacy over 4,332 persona labels. 
\emph{Acc} and \emph{Max-Ratio} are measured in \%.
%Max-Pred-ID means the label with most frequent prediction made by the attacker model and 
\emph{Acc} refers to test persona inference accuracy.
\emph{F1} uses weighted average F1-score.
\emph{Max-Ratio} indicates  the ratio that the most frequent prediction shares among all predictions.
The worse the attack model performs, the better privacy protection can be achieved.
%\yqc{I suggest we split imbalance to another table, or even just describe it in text. }
}
\vspace{-0.2in}
\end{table}

\subsection{\tbc{Privacy}}
\label{exp:4.2}
%%% attacker performance
\textbf{Attacks without Defense}.
We list the attacking performance of $\mathcal{A}$ in multiple scenarios shown in Table~\ref{tab:privacy}.
To demonstrate the overlearning issue of GPT-2, we consider 2 baseline attacks.
If the adversary has no knowledge about persona attributes distribution, then it can randomly guess over 4,332 labels  (\emph{Random Pred}).
Otherwise the adversary can perform \emph{Best Guess} that only guesses the most frequent persona in the dataset.
\emph{LM} indicates the attacker performance that only language modeling objective is applied to train the chatbot without any defense mechanism.
From the table, the test persona inference accuracy on the \emph{LM} achieves 37.59\% while guessing on the label with most occurrences merely has 0.72\% accuracy.
That is, the black-box persona inference attack has $52\times$ the accuracy of guessing.
The huge performance gap between the attacker model and the baseline guess method indicates that simple language modeling objective has serious overlearning issues that  unintentionally capture private personas of speakers.

\begin{table*}[!htbp]
\centering
\small
%  \begin{tabular}{lSSSSSSSSS}
  \begin{tabular}{lccccccccc}
    \toprule
    \multirow{2}{*}{} &
    \multicolumn{1}{c}{\multirow{2}{*}{PPL}} &
      \multicolumn{2}{c}{Distinct} &
      \multicolumn{3}{c}{BLEU} &
      \multicolumn{3}{c}{BERTScore} \\
      & & {Dist-1} & {Dist-2} & {BLEU-1} & {BLEU-2} & {BLEU-4} & {Precision} & {Recall} & {F1}\\
      \midrule
    LM & 14.821 & 0.952 & 0.879 & 0.121 & 0.0551 & 0.0123 & 0.860 & 0.843 & 0.851 \\
    %\hline
    LM+KL & 28.926 & 0.954 & 0.880 & 0.121 & 0.0558 & 0.0130 & 0.859 & 0.844 & 0.851 \\
    LM+MI & 18.740 & 0.953 & 0.880 & 0.118 & 0.0531 & 0.0121 & 0.859 & 0.843 & 0.851 \\
    LM+KL+MI & 19.674 & 0.953 & 0.880 & 0.119 & 0.0525 & 0.0105 & 0.858 & 0.842 & 0.850 \\
    \bottomrule
  \end{tabular}
  \vspace{-0.05in}
  \caption{\label{tab:utility}
Evaluation on the utility over 4,332 persona labels. 
}
\vspace{-0.05in}
\end{table*}
%%% defender performance
\textbf{Attacks on the Defensed LM}.
To avoid the persona overlearning issue, we use additional defense objectives illustrated in Section~\ref{sec:3-3}.
%\emph{LM+KL} indicates the GPT-2 is trained with language modeling and \tbc{KL} loss in Equation \ref{eqn:LM} and \ref{eqn:KL Loss 2}.
%\emph{LM+MI} applies language modeling and \tbc{MI loss} in Equation \ref{eqn:LM} and \ref{eq:final_mi} to train the GPT-2.
\emph{LM+KL+MI} utilizes language modeling, \tbc{KL loss} and \tbc{MI loss} in Equation \ref{eqn:overall loss} to train the GPT-2.
%As demonstrated in Table~\ref{tab:privacy}, all three models are able to reduce test accuracy of the black-box attacks.
As demonstrated in Table~\ref{tab:privacy}, the attacker performance on \emph{LM+KL+MI} significantly reduces the attacking accuracy from 37.59\% to 0.53\% and F1-score drops from 0.37 to nearly 0.
This defense mechanism can even outperform \emph{Best Guess} in terms of privacy protection.
That is, even if the adversary annotates its own dataset to train an attacker model, the attacking performance is still worse than simply guessing the most frequent label.
As a result, the black-box persona prediction attack becomes useless after applying the defenses for the chatbot.
The adversary cannot obtain any speaker's persona from the embedding $f(u)$ by training $\mathcal{A}$.

%\subsubsection{Analyze the Most Frequent Predictions}
To learn why the proposed defenses work so well, we further examine the ratio of the most frequent predicted label (\emph{Max-Ratio}) among all predictions.
The accuracy of \emph{Best Guess} reveals that the most frequent label in the test set has a ratio of 0.72\%.
After applying \tbc{KL loss} and \tbc{MI loss}, the attacker model tends to make predictions on a single label.
For \emph{LM+KL+MI}, the \emph{Max-Ratio} even occupies 81.87\% predictions.
This implies that the proposed defense strategies may have the potential to fool the attacker model to make wrong predictions on a single slot.
We will further investigate this implication in later sections.

Overall, the above experiment demonstrates that our proposed defense learning strategies can effectively mitigate the persona overlearning issue and avoid black-box persona inference attacks.

\begin{table*}
%\begin{adjustbox}{width=1.05\textwidth,center}
\centering
\small
  \begin{tabular}{lcccccccc}
    \toprule
    \multirow{2}{*}{} &
    %\multicolumn{1}{c}{\multirow{2}{*}{PPL($\downarrow$)}} &
      \multicolumn{4}{c}{Unseen (0-2)} &
      \multicolumn{4}{c}{Overall (0-7)} \\
%    \multirow{1}{*}{Dataset}  \\
    \cmidrule(lr){2-5}\cmidrule(lr){6-9}
    {}  & {Acc} & {F1} & {Max-Ratio}  & {$\text{BP}_u$}  & {Acc} & {F1} & {Max-Ratio}  & {$\text{BP}_u$} \\
      \midrule
    %%% add 8 clusters here
    
    Random Pred & 34.42 & 0.35 & 33.90 & 0  & 13.21 & 0.14 & 13.35 & 0   \\
    
    \tbc{Best Guess} & 56.84 & 0.41 & 100 & 2.60e-1  & 22.67  & 0.08 & 100 & 2.60e-1  \\
    %%% this one not zeroshot setting
    %\multirow{4}{*}{}  &
    %\tbc{LM} & -1 & -1 & -1 & -1  \\
    LM & 86.83 & 0.91 & 50.72 & 2.11e-3  & 72.37 & 0.72 & 20.94 & 3.04e-3\\
    LM+KL+MI & 28.68 & 0.37 & 58.15 & 2.84e-4  & 30.26 & 0.21 & 77.94 & 2.65e-4 \\
    \bottomrule
  \end{tabular}
%  \end{adjustbox}
  \vspace{-0.05in}
  \caption{\label{tab:privacy8}
Evaluation on the privacy for 8 clusters.
\emph{Unseen} shows the results only for the first 3 persona labels that defender has never seen.
\emph{Overall} refers to the results on all 8 labels.
\emph{Acc} and \emph{Max-Ratio} are measured in \%.
$\text{BP}_u$ corresponds to Bayesian Privacy loss on the uniform distribution.
%Max-Pred-ID means the label with most frequent prediction made by the attacker model and 
%\emph{Acc} refers to test persona inference accuracy.
%\emph{F1} uses weighted average F1-score for 4332 persona labels.
%\emph{Max-Ratio} indicates  the ratio that the most frequent prediction shares among all predictions.
Still, the worse the attack model performs, the better privacy protection can be achieved. 
}
\vspace{-0.2in}
\end{table*}

\subsection{Ablation Study}
\label{exp:ablation}
To show the effectiveness of proposed \tbc{KL loss and MI loss} and how they affect the performance of black-box persona inference attacks, we consider the inclusion and exclusion of proposed defense objectives.
The result is shown in Table~\ref{tab:privacy}.
\emph{LM+KL} indicates the GPT-2 is trained with language modeling and \tbc{KL} loss.
%in Equation \ref{eqn:LM} and \ref{eqn:KL Loss 2}.
\emph{LM+MI} applies language modeling and \tbc{MI loss}.
%in Equation \ref{eqn:LM} and \ref{eq:final_mi} to train the GPT-2.
From the table, it can be seen that \emph{LM+KL}, \emph{LM+MI} and \emph{LM+KL+MI} are all able to reduce the test accuracy of the attacks.
The \tbc{KL loss} is weaker from the perspective of defense, but it tends to flatten the estimated persona distribution with much smaller \emph{Max-Ratio}.
The \emph{LM+MI} shares similar test accuracy and F1-score with \emph{LM+KL+MI}, but nearly all predictions are made on a single persona label with a ratio of 99.84\%.
This suggests that \tbc{MI loss} causes the attacker model to predict all labels on a single persona attribute.
After \tbc{KL loss} is applied on \emph{LM+KL+MI}, the \emph{Max-Ratio} drops to 81.87\%.
%%% Not sure to add

As discussed earlier, high \emph{Max-Ratio} may also cause privacy leakage.
Suppose the adversary knows the persona with \emph{Max-Ratio}, then it can improve its guess by not predicting this persona, which is a threat for fewer persona labels (for example, binary classification).
These results verify that \tbc{KL loss} introduces flatter estimation and \tbc{MI loss} is more effective against persona overlearning, which conforms to our intuition of their objectives in Section \ref{sec:3-3}.

\subsection{Utility}
\label{exp:4.3}
Besides privacy,  utility is another key objective to train a chatbot.
Several automatic metrics are considered to evaluate the generation performance.
For generation, we use GPT-2 to generate responses of the second speaker (\emph{Human B} in Figure \ref{fig:atk-full-sample}) with all previous turns as context.
Then we compared the generated model outputs with ground truth replies.
We use \emph{Dist-1} and \emph{Dist-2} to count ratios of distinct unigrams and bigrams.
\emph{BLEU-1}, \emph{BLEU-2} and \emph{BLEU-4} are applied to evaluate generation similarity with ground truth.
Due to the one-to-many nature of chit-chats, the \emph{BLEU} is not adequate to compare generated responses with ground truth.
Hence, we adapt \emph{Precision}, \emph{Recall} and \emph{Precision} of \emph{BERTScore} to measure the similarity in the embedding space.

The evaluation result is shown in Table  \ref{tab:utility}, where same models from Table \ref{tab:privacy} are evaluated.
%\emph{LM+KL+MI} utilizes language model, \tbc{KL loss} and \tbc{MI loss}.
The result indicates that adding \tbc{KL loss} will increase the perplexity greatly from 14.8 to 28.9.
After combining \tbc{KL loss} with \tbc{MI loss}, its perplexity decreases to 19.674.
%%% talk more about MI, need exp result later
A plausible explanation is that \tbc{KL loss} confuses the persona predictor and indirectly increases the uncertainty of the GPT-2.
All GPT-2 models have relatively low \emph{BLEU} scores due to the one-to-many mapping between contexts and responses. 
For \emph{Distinct} and \emph{BERTScore}, there are only minor differences between  LM and defensed LMs.
Though the uncertainty increases after applying \tbc{KL loss} and \tbc{MI loss}, it does no harm to the quality of generation.
In summary, there is almost no negative influence on the utility after applying the proposed defense strategies.

\begin{figure*}[t]
\centering
\makebox[0pt]{
\includegraphics[width=1\textwidth]{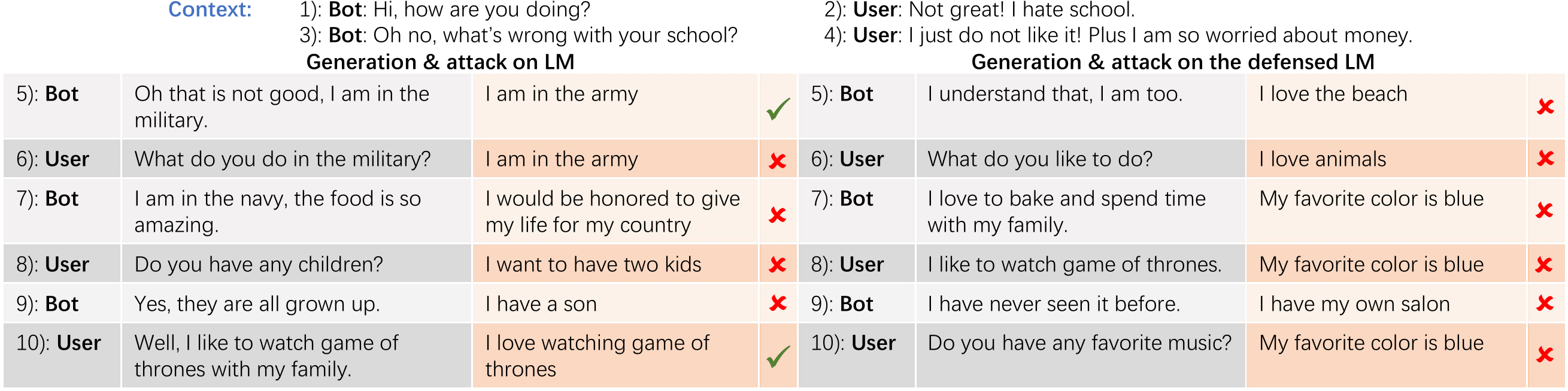}
}
\vspace{-0.1in}
\caption{
Black-box persona inference attacks on chit-chats between the user (the authors of this paper) and chatbots.
For both conversations, the ``context'' is fixed and used as the first four utterances. 
Then the bot and the user start interactive conversations with the ``context''. 
Since there is no gold standard, the results are annotated by the authors.
%\yqc{In the caption, you need to explain what is ``context,'' ``how do you generate the responses,'' and  ``how do you evaluate the attack defense results.'' E.g., since there is no gold standard, the results are annotated by the authors of this paper.}
}
\label{fig:case_study}
\vspace{-0.1in}
\end{figure*}
\subsection{More Setups on Attacks}
\label{exp:More Setups}
\textbf{Attacks on Imbalanced Data Distribution}.
%\label{exp: imbalance}
Previous black-box attacks usually assume that the annotated dataset $D_a$ must share similar data distribution with the defender's training data.
To examine the performance of defense strategies on unseen personas, we assign the adversary's dataset $D_a$ with labels that the defender cannot acquire.
We split data with 500 persona labels that are uniquely held by the adversary.
%\yqc{We need to provide more details here: how many training and testing data are there? How many training categories/testing categories? Please say this explicitly. Reviewers won't do counting.}
The defender owns 8,031 conversations with persona labels ranging from 500 to 4,331 while the adversary holds 2,376 dialogues with persona labels ranging from 0 to 4,331.
For testing, 500 conversations with persona labels ranging from 0 to 4,331 are used.
%\emph{Imbalance} of Table \ref{tab:privacy} shows the attacker performance on the imbalanced data distribution setup.

Under imbalanced data distribution, the attack on the defensed LM has \emph{Acc} 0.47\%, \emph{F1} 1.90e-3 and \emph{Max-Ratio} 94.06\%.
The persona inference accuracy is still very low and the attacker model tends to predict more on a single persona label than the balanced data distribution setup.
This result shows that the proposed overall loss can also prevent black-box persona inference attacks on unseen personas.
It also verifies previous suggestions that combining \tbc{LM loss} with \tbc{MI loss} may fool the attacker model to make wrong predictions.

\textbf{Attacks on Fewer Persona Labels}.
The above experiments are based on 4,332 persona labels.
In fact, many personas share similar meanings and can be further clustered.
Besides, to better evaluate privacy loss for the estimated distribution, a smaller label space is preferred. 
%Besides, to evaluate privacy loss for the estimated distribution with Bayesian Privacy (BP)~\cite{gu-2021-federated}, a smaller label space is preferred. 
%since the distribution of a large $C$ is too sparse.
Therefore, it is necessary to consider defense performance on a smaller label space.
We use Sentence-BERT ~\cite{reimers-2020-multilingual-sentence-bert} to embed all persona sentences and perform k-means clustering on the embeddings to obtain 8 clusters.
We manually checked these clusters and classified them as 
cars,
food,
animals (pets),
family information,
hobbies,
jobs,
personal information
and music tastes respectively.
%%%%%%%%%% hr added for clustering result explanation
To evaluate how the clustering performs, we randomly sample 100 utterances with clustered labels and invite two volunteers to inspect those samples. Both of them agree on 90\% of the clustered annotations.
After manual inspection of the remaining 10\% annotations, the clustering error rate is 8\%.
%and we fix 5\% of them.
Following previous imbalanced data split, 
%in Section~\ref{exp: imbalance}, 
we assign data in the first 3 clusters only to the adversary to make the data distribution imbalanced. 
%\yqc{Again, here is not very clear. Please say training/testing numbers for defender and attacker explicitly.}
Here, the defender owns 6,654 conversations with persona labels ranging from 3 to 7 while the adversary holds 3,753 dialogues with persona labels ranging from 0 to 7.
For testing, 500 conversations with persona labels ranging from 0 to 7 are used.

The attacking performance for both unseen labels and all labels is displayed in Table \ref{tab:privacy8}.
$\text{BP}_u$ measures the KL divergence $D_{KL}(F_0 || \mathcal{A}(f(u)))$ where $F_0$ refers to uniform distribution.
%or the true data distribution ($\text{BP}_d$).
%$\text{BP}_u$ indicates Bayesian Privacy is estimated on the uniform distribution and  $\text{BP}_p$ refers to Bayesian Privacy with true data distribution.
For imbalanced data distribution with a small label space, our proposed defenses can still achieve much lower attack accuracy than \emph{LM} on both \emph{Unseen} and \emph{Overall}.
However, for \emph{Overall}, \emph{LM+KL+MI} has higher accuracy with a lower F1-score compared with two baselines.
%%% careful
This indicates that proposed defenses fail to protect privacy as we desired in the baselines.
%%% BP part
For $\text{BP}_u$, \emph{LM+KL+MI} are around 10 times smaller than \emph{LM}.
It means that after applying defense objectives, the attacker's estimated distribution is much closer to the uniform distribution.
Thus the effectiveness of the KL loss is verified.
%For $\text{BP}_d$, the performance of \emph{LM+KL+MI} and \emph{LM} are similar.
%Moreover, their $\text{BP}_d$ are both smaller than \emph{Random Pred}.
%Smaller $\text{BP}_d$ shows that they can both estimate the true distribution better than random guesses.
In addition, \emph{Max-Ratio} with 8 clusters on \emph{Unseen} is smaller than 4,332 labels even though the distribution of 8 clusters is obviously tighter.
Still, the \emph{Max-Ratio} of 58.15\% accounts for a much larger fraction than other predictions. 
In summary, the above results imply that for the smaller label space, our proposed defense objectives are still effective even on unseen persona labels.

%fig:case_study
\subsection{Case Study}
\label{case}
In Figure \ref{fig:case_study}, we give an example of the persona inference attack, where  conversations are generated between the chatbot and the user with the given context.
We manually mark True/False on the predicted results.
As shown in the figure, there are several successful attacks on LM and no correct prediction on the defensed LM.
For attacks on LM, speakers' hobbies and jobs can be inferred.
For incorrect predictions, the attacker model can still predict context-aware personas.
%the adversary can learn that the \emph{Bot} serves in the army and the \emph{User} likes Game of Thrones.
After applying proposed defense learning strategies, the predicted personas become irrelevant with context and mostly predict ``My favorite color is blue.''
In fact, it is the most frequent prediction for  \emph{LM+KL+MI} over 4,332 persona labels.
This attack example illustrates that our defense objectives can prevent the black-box persona inference attack from inferring relevant personas.

\begin{comment}
\subsection{Case Study}
In Figure \ref{fig:atk full sample}, we give an example on the persona inference attack. 
Here, attacker models without defense (\emph{LM}) and with defence (\emph{LM+KL+MI}) try to infer personas of \emph{Human A} and \emph{Human B} for all utterances.
Since some ground truth personas are missing, we manually mark True/False on the predicted results.
As shown in the figure, there are several successful attacks on \emph{LM} and no correct prediction on \emph{LM+KL+MI}.
For successful attacks, speakers' hobbies and jobs can be inferred.
For incorrect predictions, the attacker model can still predict context-aware personas without proper defenses (\emph{LM}).
After applying proposed defense learning strategies, the predicted personas become irrelevant with context and mostly predict ``My favorite color is blue''.
In fact, ``My favorite color is blue'' is the most frequent prediction for  \emph{LM+KL+MI} over 4,332 persona labels.
This attack example illustrates that our defense objectives can prevent the black-box persona inference attack from inferring relevant personas.
\end{comment}

%Besides only considering attack results with incomplete annotated labels, we give another example that attacks all utterances in Figure \ref{fig:atk full sample}.
%Here we manually mark True/False on the predicted results.
%Still, it is shown that the attacker model can predict context-aware personas without proper defenses (\emph{LM}).
%After applying proposed defense objectives, the attacker model can no longer predict any relevant personas.
\section{Conclusion}
In this paper, we show that LM-based chatbots tend to reveal personas of speakers and propose effective defense objectives to prevent GPT-2 from overlearning.
Unlike other works that suffer from utility degradation, our defense learning strategies do no harm to the powerful generation ability of LM-based chatbots.
We conduct extensive experiments to evaluate both privacy and utility.
We perform black-box persona inference attacks under various setups to demonstrate the robustness of proposed defense learning strategies.
In addition, we use automatic metrics to show that proposed defense learning strategies maintain the utility.
For future work, we suggest working on flattening the distributions of attacker models.

\section{Ethical Considerations}
We declare that all authors of this paper acknowledge the \emph{ACM Code of Ethics} and honor the code of conduct.
This work essentially considers black-box attacks on the private persona attributes and proposes effective learning strategies to prevent chatbots from overlearning private personas.

\textbf{Dataset}.
During our dataset collection, all the conversations and personas are collected from publicly available datasets including PersonaChat and DNLI.
All the speakers are anonymized and no identifiable personal information is included.
%Hence, no privacy is violated for both datasets and our collected dataset.

\textbf{Model}.
For training our LM-based chatbots, we follow standard methods.
We are well aware of the bias issue inside current language models.
In the future, if there are other fairer language models, we will extend our defenses on them.

\section*{Acknowledgment}
The authors of this paper were supported by the NSFC Fund (U20B2053) from the NSFC of China, the RIF (R6020-19 and R6021-20) and the GRF (16211520) from RGC of Hong Kong, the MHKJFS (MHP/001/19) from ITC of Hong Kong and the National Key R\&D Program of China (2019YFE0198200) with special thanks to Hong Kong Mediation and Arbitration Centre (HKMAAC) and California University, School of Business Law \& Technology (CUSBLT), and the Jiangsu Province Science and Technology Collaboration Fund (BZ2021065).

%\section*{Acknowledgements}

\clearpage
% Entries for the entire Anthology, followed by custom entries
\bibliography{anthology,custom}
\bibliographystyle{acl_natbib}

%\afterpage{\clearpage}
\clearpage
\appendix
\begin{figure}[t]
\centering
\includegraphics[width=1.03\columnwidth]{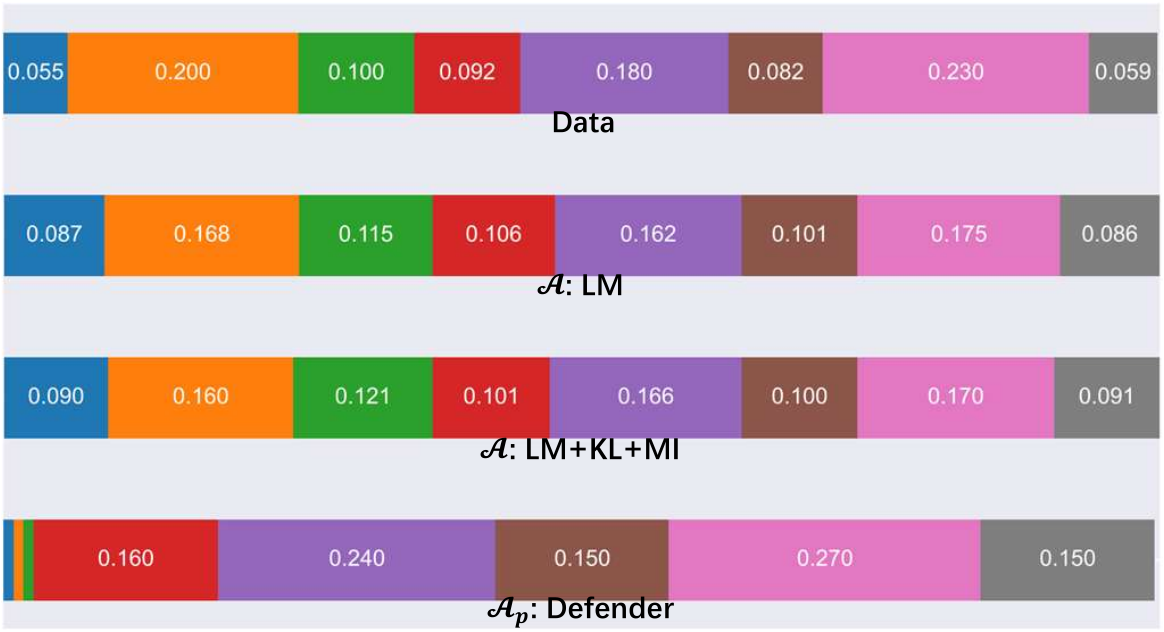}
\caption{The test set data distribution (\emph{Data}) and average estimated distributions of persona predictors (\emph{$\mathcal{A}$} and \emph{$\mathcal{A}_p$}) over imbalanced 8 clusters.
}
\label{fig:logits}
\end{figure}

\begin{comment}
\begin{table*}[htbp]
\centering
  \begin{tabular}{lcccccccccc}
    \toprule
%    \multirow{1}{*}{Dataset}  \\
    {}  & {0} & {1} & {2} & {3}  & {4} & {5} & {6} & {7} & {$\text{KL}_{\text{data}}$}  &  {$\text{KL}_{\text{uni}}$}\\
      \midrule
      
    Data & 0.055 & 0.20 & 0.10 & 0.092 & 0.18 & 0.082 & 0.23 & 0.059 & 0 & 0.13\\
    Attacker & 0.061 & 0.23 & 0.16 & 0.071 & 0.17 & 0.068 & 0.18 & 0.055 & 0.025 & 0.14\\
    Fake Attacker & 0.0085 & 0.0087 & 0.0084 & 0.16 & 0.24 & 0.15 & 0.27 & 0.14 & 0.74 & 0.035  \\
    \bottomrule
  \end{tabular}
  \caption{\label{tab:logits}
The test set distribution and average estimated distribution of persona predictors over 8 clusters.
\emph{$\text{KL}_{\text{data}}$} refers to the KL divergence with the \emph{Data} distribution and \emph{$\text{KL}_{\text{uni}}$} indicates the KL divergence with uniform distribution.
% Notice that \emph{Fake Attacker} only has access to last 5 labels and its \emph{$\text{KL}_{\text{uni}}$} is calculated only based on these 5 labels while \emph{$\text{KL}_{\text{uni}}$} of \emph{Attacker} considers all 8 labels.
}
\end{table*}
\end{comment}

\section{Training details.}
For each conversation, the utterances are concatenated by the special token  ``<|endoftext|>'' to train the GPT-2.
To decode outputs from GPT-2, we apply the Nucleus Sampling~\cite{Holtzman-2020-TheCC} method.
We set top-p = 0.9 with a temperature coefficient 0.9 to sample words from the GPT-2.
For optimization, we set 2 AdamW optimizers ~\cite{Loshchilov-2018-decoupled} for the chatbot and the persona predictor respectively.
The learning rate is 3e-5 with linear warm-up and decay.
For hyper-parameters, we set $\lambda_0 = 1$, $\lambda_1 =10$ and $\lambda_2 = 1$.

\section{Comparison of Internal Distribution between $\mathcal{A}$ and $\mathcal{A}_p$}
\label{exp:logits}

To make predictions on personas, the $\arg\max$ function is used for the estimated distribution of persona predictors.
However, the internal distribution conveys crucial information  about how the persona predictors estimate $f(u)$.
We follow the setup of imbalanced data split of 8 clusters in Section \ref{exp:More Setups} to examine persona predictors of attacker $\mathcal{A}$ and fake attacker $\mathcal{A}_p$.

Figure \ref{fig:logits} shows the data distribution of the test set and average distribution after softmax activation over the 8 labels for attacker $\mathcal{A}$ and defender $\mathcal{A}_p$.
%Moreover, we also calculate their KL divergence with ground truth distribution and uniform distribution.
For attacker $\mathcal{A}$, we consider the attack on \emph{LM} and \emph{LM+KL+MI}.
The defender $\mathcal{A}_p$ tends to have a large difference with \emph{Data} and tries to flatten its distribution among its own training set (the last 5 labels).
This behavior conforms to the KL loss's objective that aims to flatten the distribution and deviate from the ground truth distribution.
For attacker $\mathcal{A}$, distributions of both \emph{LM} and \emph{LM+KL+MI} seem close to the ground truth distribution.
This indicates that the attacker model $\mathcal{A}$ can still learn statistical information about personas.
However, its attacking performance is poor.
%in terms of \emph{Imbalance} from Table \ref{tab:privacy}.
The poor performance implies our proposed defense learning strategies may obfuscate \emph{Attacker} for estimating single sample $f(u)$ and finally make the wrong prediction.

\begin{table*}
%\begin{adjustbox}{width=1.05\textwidth,center}
\centering
\small
  \begin{tabular}{lcccccccc}
    \toprule
    {}  & {Top-1} & {Top-3} & {Top-5}  & {Top-10}  & {Top-50} & {Top-100} & {Top-500}  & {Top-2000} \\
      \midrule
    %%% add 8 clusters here
    
    LM & 37.59 & 55.57 & 63.28 & 72.76  & 87.19 & 91.54 & 97.79 & 99.60\\
    LM+KL+MI & 0.53 & 1.80 & 2.24 & 3.20 & 8.64 & 12.10 & 30.57 & 80.22 \\
    \bottomrule
  \end{tabular}
%  \end{adjustbox}
  \vspace{-0.07in}
  \caption{\label{tab:topk}
Evaluation on the top-k accuracy over 4,332 clusters.
}
\vspace{-0.1in}
\end{table*}
\begin{figure}[t]
\centering
\includegraphics[width=1\columnwidth]{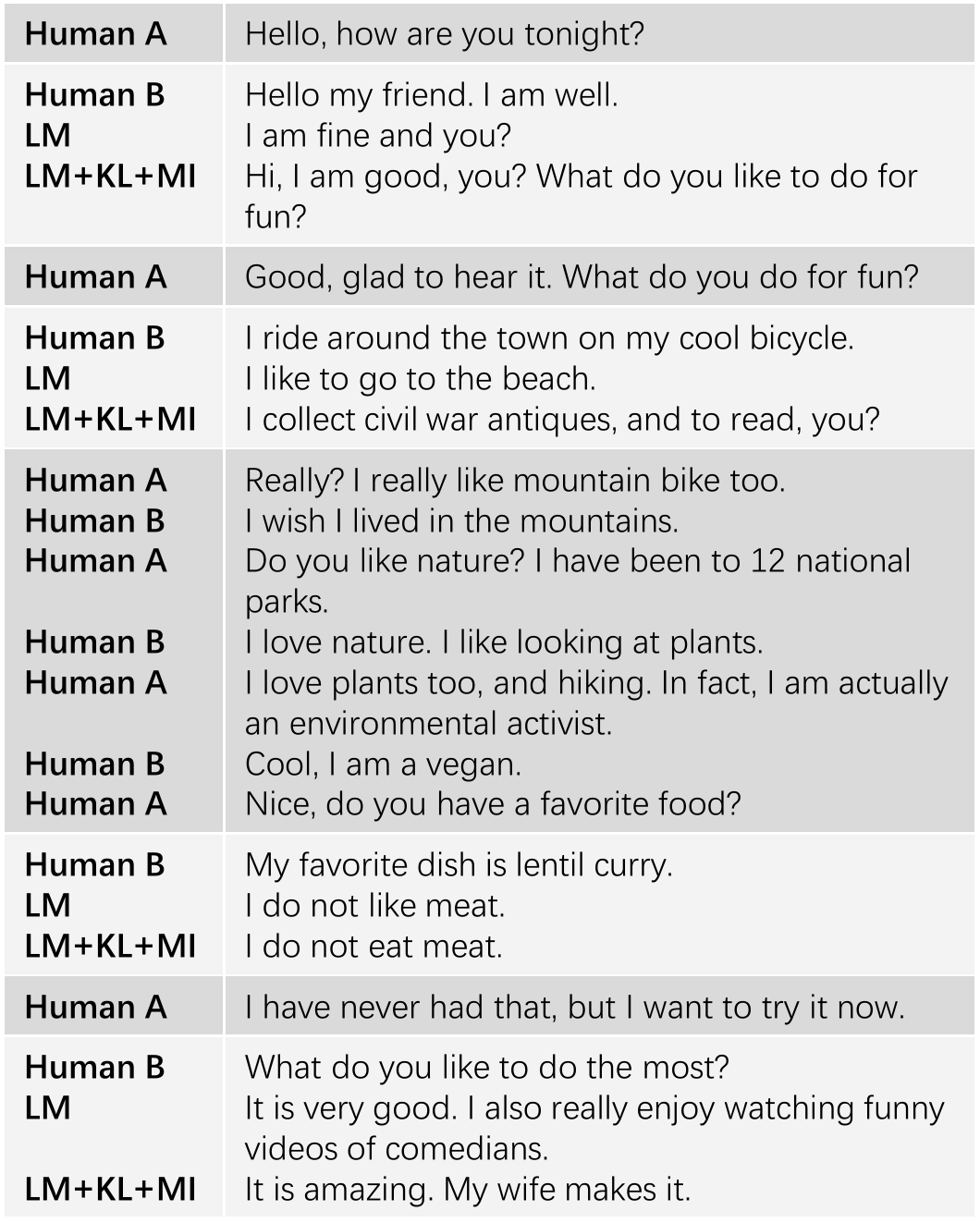}
\caption{Dialog generation example on \emph{Human B}. All previous utterances between A and B are used as context to generate responses.}
\label{tab:dialog sample}
\end{figure}

\begin{table}[t]
\small
\begin{tabular}{p{0.95\linewidth}}%{l}
\toprule
\textbf{Dialog Context 1} \\
\textbf{Human A}: Hi how are you doing?\\
\textbf{Human B}: I am great and you?\\
\textbf{Human A}: I am great, just reading.\\
\textbf{Human B}: I am listening to the rolling stones, I love them.\\
\textbf{Human A}: Is that your favorite band? \\
\textbf{Human B}: Yes it is. I am working right now too. \\
\textbf{Human A}: Where do you work at? \\
\textbf{Human B}: \textcolor{orange}{IBM in Chicago, what about you?} \\
\hline
\textbf{Persona Prediction}: \\
\textbf{LM}: I currently work for IBM in Chicago. \textcolor{green}{\textbf{\cmark}}\\
\textbf{LM+KL+MI}: I love cats. \textcolor{red}{\textbf{\xmark}}\\

\midrule

\textbf{Dialog Context 2} \\
\textbf{Human A}: Hello there my name is Dr.Lucy. How are you?\\
\textbf{Human B}: I am great, loving this city life, how are you?\\
\textbf{Human A}: I am well thank you. I miss my country life in Spain.\\
\textbf{Human B}: My older brother lives in Spain, how is it?\\
\textbf{Human A}: It is beautiful. I hope to take my family back there.\\
\textbf{Human B}: Yes, maybe i will take my girlfriend that I love there one day.\\
\textbf{Human A}: Oh, how long have you two been together? \\
\textbf{Human B}: Very long, she was with me when I colored my hair pink. \\
\textbf{Human A}: That is awesome. What type of music do you two listen to? \\
\textbf{Human B}: I like reading music, what about you?\\
\textbf{Human A}:  \textcolor{orange}{Hip hop is my favorite. Do you play an instrument?} \\
\hline
\textbf{Persona Prediction}: \\
\textbf{Ground truth}: My favorite music is hip hop. \\
\textbf{LM}: I know how to play the guitar. \textcolor{red}{\textbf{\xmark}}\\
%my parents died in a plane crash
\textbf{LM+KL+MI}: My favorite food is pizza. \textcolor{red}{\textbf{\xmark}}\\
\bottomrule
\end{tabular}
%Dialog Context \\
%\hline
%Speaker A: \\
\caption{More persona inference attack examples.
The embeddings of the final utterance with orange color are used for inferring B's persona. 
}
\label{app-tab:inference sample}
\end{table}

\section{More on Case Study}
\label{exp:case study}

\subsection{Example of Generation}
\label{exp:sample example}
To show an intuition view on utility, we provide one generation sample shown in Figure \ref{tab:dialog sample}.
Both \emph{LM} and \emph{LM+KL+MI} are able to generate fluent and proper relies.
Moreover, they tend to maintain coherence with previous contexts.
For example, it is mentioned in the context that \emph{Human B} is a vegan and both chatbots respond  that they do not eat meat for the food preference.
This generation example shows that proposed defense learning objectives preserve the model utility.

\subsection{More Examples of Persona Inference}
%In Table \ref{tab:inference sample}, we give an example on the persona inference attack. 
%Here, attacker models without defense (\emph{LM}) and with defence (\emph{LM+KL+MI}) try to infer \emph{Human B}'s persona through the embeddings of the orange text.
%The prediction result on \emph{LM} show that B's sensitive family information can be inferred from embeddings of B's utterances.
%After applying proposed defense learning strategies, the prediction result becomes irrelevant with context.
%In fact, ``My favorite color is blue.'' is the most frequent prediction for  \emph{LM+KL+MI} over 4,332 persona labels.
%This attack example illustrates that our defense objectives can prevent the black-box persona inference attack from inferring relevant personas.

%\section{More Examples}
Here, we give two more examples of the persona inference attacks in Table \ref{app-tab:inference sample}.
The first example shows one successful defense.
For the second example, both attackers with and without defense fail to predict the ground truth persona.
Still, we can see that \emph{LM+KL+MI} predicts personas that are irrelevant to the context.
However, \emph{LM}'s output ``I know how to play the guitar.'' is much closer to the context about music and instruments.
Without any defense, the above examples show that the attacker model can still predict context-aware personas even if its predictions are wrong.
After applying the proposed defenses, the attacker model cannot predict meaningful personas relevant to the context.

%Besides only considering attack results with incomplete annotated labels, we give another example that attacks all utterances in Figure \ref{fig:atk full sample}.
%Here we manually mark True/False on the predicted results.
%Still, it is shown that the attacker model can predict context-aware personas without proper defenses (\emph{LM}).
%After applying proposed defense objectives, the attacker model can no longer predict any relevant personas.

\section{Evaluation on Top-k Accuracy}
Previous experiments mainly consider accuracy as the evaluation metric.
In this section, we use top-k accuracy for the black-box persona inference attacks to measure  privacy protection.
As shown in Table \ref{tab:topk}, our defense is much more robust than \emph{LM} when $\text{k} \leq 50$.
When k is larger than 500, the defense degrades rapidly as k increases.
This result implies that the ground truth personas mostly lie in the top 2,000 predictions even if the defense is applied.
For a smaller k, our proposed defense learning strategies are still effective.
%\section{Example Appendix}
%\label{sec:appendix}

%This is an appendix.
\end{document}